\documentclass[letterpaper]{article} 
\usepackage{aaai2027}  
\usepackage[hyphens]{url}  
\usepackage{graphicx} 
\usepackage{natbib}  
\usepackage{caption} 
\usepackage{algorithm}
\usepackage{algorithmic}
\usepackage{amsmath}
\usepackage{newfloat}
\usepackage{listings}
\DeclareCaptionStyle{ruled}{labelfont=normalfont,labelsep=colon,strut=off} 
\floatstyle{ruled}
\newfloat{listing}{tb}{lst}{}
\floatname{listing}{Listing}

\usepackage{booktabs}

\newcommand{\method}{FOVEA}
\newcommand{\vlm}{VLM}
\newcommand{\kv}{KV}
\usepackage{multirow}
\usepackage{amsfonts}
\newcommand{\sr}[1]{\ensuremath{#1\!\times}}
\newcommand{\best}[1]{\ensuremath{\mathbf{#1\!\times}}}
\newcommand{\second}[1]{\ensuremath{\underline{#1\!\times}}}
\newcommand{\secondval}[1]{\ensuremath{\underline{#1}}}
\usepackage{pifont}
\usepackage{xcolor}
\usepackage{colortbl}

\definecolor{algcommentgreen}{RGB}{0,125,0}

\newcommand{\AlgComment}[1]{%
    \item[]\textcolor{algcommentgreen}{%
        $\triangleright$~#1%
    }%
}

\definecolor{methodgroupgray}{RGB}{242,243,245}
\newcommand{\methodgroup}[1]{%
  & \multicolumn{12}{c}{\cellcolor{methodgroupgray}\textit{\textbf{#1}}}\\
}
\newcommand{\arbaseline}{%
  & AR Baseline
  & \sr{1.00} & \sr{1.00} & \sr{1.00} & \sr{1.00} & \sr{1.00}
  & \sr{1.00} & \sr{1.00} & \sr{1.00} & \sr{1.00} & \sr{1.00} & -- \\
}

\title{\textbf{\method: Focused On-Demand Visual Evidence Adaptation for Cache-Friendly Multimodal Speculative Decoding}}

\author{
    Hengjie Zhu\textsuperscript{\rm 1,2},
    Dayan Wu\textsuperscript{\rm 1}\corresponding,
    Zihao Zhang\textsuperscript{\rm 1,2},
    Xinze Liu\textsuperscript{\rm 1,2},\\
    Jingxuan Yu\textsuperscript{\rm 3},
    Peng Fu\textsuperscript{\rm 1},
    Zheng Lin\textsuperscript{\rm 1},
    Weiping Wang\textsuperscript{\rm 1},
    Ding Wang\textsuperscript{\rm 1}
}

\affiliations{
    \textsuperscript{\rm 1}Institute of Information Engineering, Chinese Academy of Sciences\\
    \textsuperscript{\rm 2}School of Cyber Security, University of Chinese Academy of Sciences\\
    \textsuperscript{\rm 3}School of Cyber Science and Engineering, Southeast University\\
    \{zhuhengjie, wudayan, zhangzihao, liuxinze, fupeng, linzheng, wangweiping, wangding\}@iie.ac.cn,
    yujingxuan24@seu.edu.cn
}

\begin{document}

\maketitle

\begin{abstract}
Multimodal speculative decoding accelerates vision-language models by allowing a lightweight draft model to propose candidate tokens for parallel verification by a larger target model.
Existing methods typically condition the drafter on a fixed visual interface, such as a predefined visual-token budget or a static compressed representation.
However, our controlled visual-budget analysis shows that visual demand varies substantially across tasks and decoding stages, which means more visual input is not always beneficial.
Actually, insufficient evidence may weaken visual grounding, while excessive context adds overhead and may disrupt drafting.
We propose \method{} (Focused On-demand Visual Evidence Adaptation), a
cache-friendly approach that builds a reusable visual memory and dynamically
retrieves a bounded subset for a draft state. A cumulative-mass rule determines both how many and which entries are selected.
The selected entries are aggregated into a visual readout and fused with the
current draft hidden state through a lightweight gated residual correction.
Rather than inserting visual tokens into the autoregressive context, the
correction modifies only the representation passed to the language-model head.
Experiments across multiple vision-language backbones and multimodal
benchmarks show that \method{} improves draft acceptance and
end-to-end decoding speed, achieving up to $2.13\times$ speedup over
autoregressive decoding. These results demonstrate that state-conditioned evidence retrieval is an effective alternative to reusing a fixed visual representation throughout multimodal generation.
\end{abstract}

\section{Introduction}
\begin{figure}[!t]
    \centering
    \includegraphics[width=0.96\columnwidth]{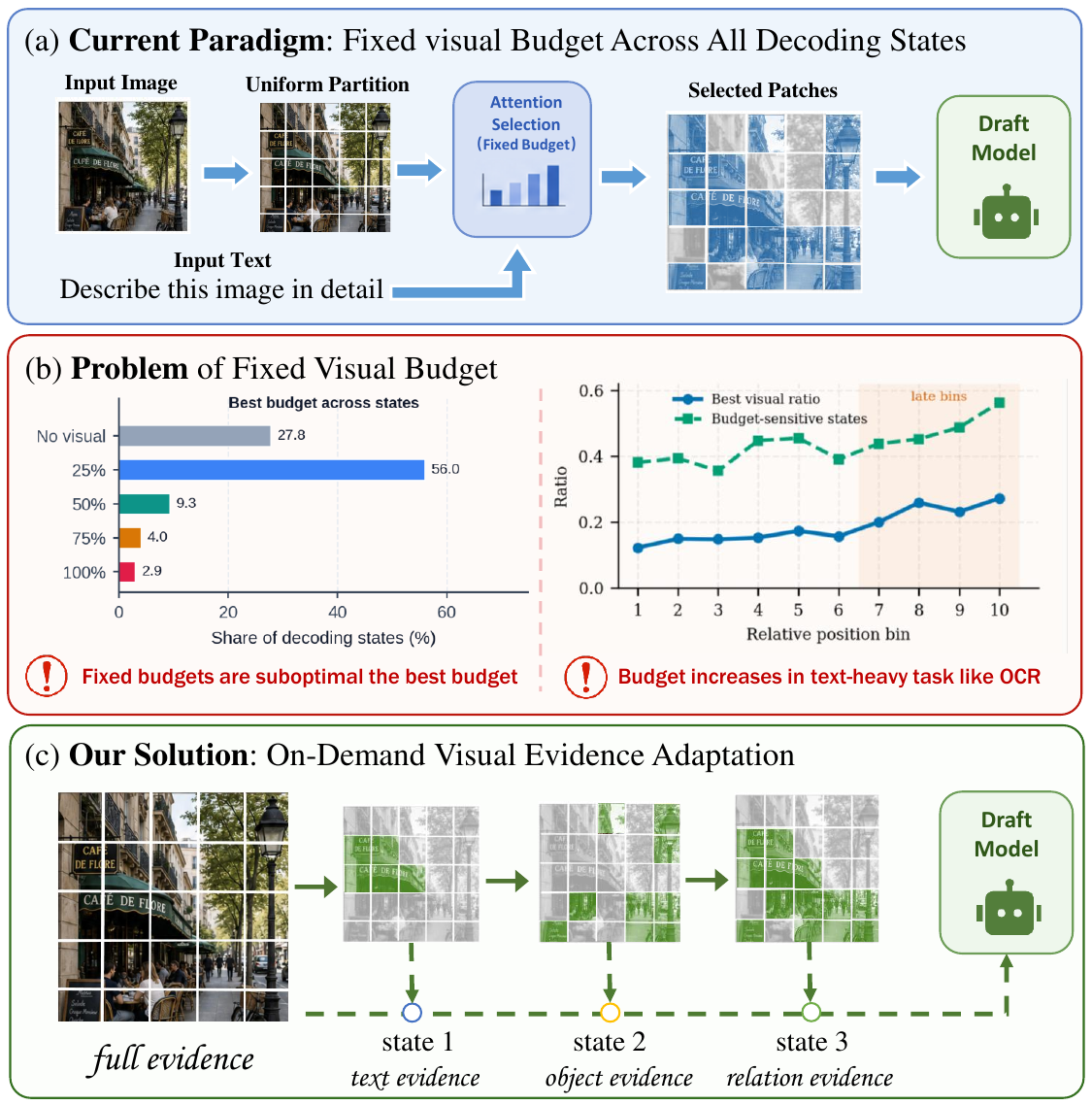}
    \caption{
    Our visual-budget diagnostics in (b) show that the preferred diagnostic budget varies across decoding states and tends to be higher later in text-heavy generation.
    These findings expose the limitation of the fixed-budget interface in (a), which provides the same visual budget to every state.
    To address the content-selection aspect of this mismatch, (c) \method{} stores visual evidence once and retrieves a dynamic number of state-conditioned entries, allowing different draft states to select text-, object-, or relation-centric information without modifying the cached token context.
    }
    \label{fig:intro}
\end{figure}

Speculative decoding \cite{chen2023accelerating, leviathan2023fast, ku2024findings} accelerates autoregressive generation by dividing computation between a lightweight draft model and a larger target model \cite{berdoz2026steering, do2026adaspec, shi2026scaling}.
The draft model expands a candidate sequence or tree, and the target model verifies several candidate tokens in one forward pass.
For text-only language models \cite{achiam2023gpt, touvron2023llama}, this separation has produced practical acceleration with methods such as speculative decoding \cite{leviathan2023fast}, Medusa \cite{cai2024medusa}, EAGLE series ~\citep{li2024eagle,li2024eagle2,li2026eagle}, and DFlash \cite{chen2026dflash}.
However, for multimodal large language models \cite{li2023blip, liu2023visual, lu2022learn}, drafting introduces an additional perceptual challenge.
A visually grounded token may depend on a small object, OCR text, spatial relation, or chart element among many projected visual tokens.
The draft model must therefore be cheap enough to accelerate decoding while still receiving the image evidence needed by the current prediction.

Most multimodal draft methods address this tension through a fixed visual interface, either by exposing the drafter to many visual tokens~\citep{huang2025specvlm} or by reusing a compressed visual representation throughout generation~\citep{hu2026dream,kang2026vispec,liu2026dream}.
Despite their implementation differences, these methods assume that \textbf{different samples and decoding states require a similar amount or form of visual evidence.}
Our visual-budget diagnostic challenges this assumption.
Across visual-budget settings ranging from no visual input to full visual access, the preferred budget varies substantially both within the same image and across decoding states.
As shown in Figure~\ref{fig:intro}(b), 25\% budget performs best for 56.0\% of the evaluated states, whereas high-visual settings such as 75\% and full access are rarely optimal.
Moreover, visual demand changes across draft branches and generation stages, especially in text-heavy tasks where later predictions may require fine-grained textual evidence.
These findings show that no single global visual interface is uniformly suitable and motivate selecting evidence according to the current draft state.

Supporting such state-conditioned evidence selection raises two challenges.
First, \textbf{the drafter must determine which visual evidence is relevant to the current state and retrieve it with little overhead, without repeatedly processing the full visual sequence.}
Second, \textbf{state-conditioned selection must remain compatible with \kv{}-cache reuse.}
Dynamically inserting or removing selected visual tokens would change the autoregressive context and invalidate previously cached hidden states and \kv{} entries.
Recomputing the cache would reduce speculative speedup, whereas reusing an inconsistent cache could impair drafting.
The selected visual evidence should therefore influence current-token scoring without altering the historical context or its associated cache.

To address these challenges, we propose \method{} (Focused On-demand Visual Evidence Adaptation), a dynamic-cardinality, state-conditioned visual-evidence selection mechanism for multimodal speculative decoding.
After the image is encoded by the target model, \method{} projects the resulting visual tokens once to construct a reusable image-specific memory.
As shown in Figure~\ref{fig:intro}(c), during draft-tree decoding, a draft hidden state queries this memory and selects the top entries relevant to the current decoding state.
Rather than inserting the retrieved evidence into the draft model's cached context, \method{} fuses it through a lightweight gated residual module.
Thus, \method{} adapts the representation used to score each draft node without modifying the autoregressive token context or invalidating the historical \kv{} cache.

Our contributions are as follows:
\begin{itemize}
    \item We identify state-varying and non-monotonic visual-budget sensitivity in multimodal speculative decoding, showing that one globally fixed visual interface is not uniformly suitable across branches and generation stages.
    \item We propose \method{}, which uses a dynamic retrieval cardinality to select state-conditioned evidence from a reusable visual memory and applies it through gated hidden-state correction without invalidating the historical \kv{} cache.
    \item Extensive experiments across three vision-language backbones and nine multimodal benchmarks show that \method{} improves draft acceptance and end-to-end decoding speed over strong baselines, achieving up to $2.13\times$ speedup over autoregressive decoding.
\end{itemize}

\section{Related Work}

\paragraph{Multimodal speculative decoding}
Speculative decoding accelerates autoregressive generation by letting a lightweight drafter propose candidate tokens that a larger target model verifies in parallel~\citep{leviathan2023fast}.
For text-only language models, this draft--verify paradigm has been improved with auxiliary decoding heads, feature-level drafting, and dynamic draft trees~\citep{cai2024medusa,li2024eagle,li2024eagle2,li2026eagle}.
For vision-language models \cite{wang2024qwen2, wang2024emu3}, the drafter must approximate both language continuation and visual grounding, making acceptance behavior sensitive to visual conditioning~\citep{gagrani2024speculative}.

Recent multimodal speculative decoders improve drafting through vision-aware adaptation~\citep{kang2026vispec}, target-feature injection~\citep{hu2026dream}, multimodal adaptation or distillation~\citep{ganesan2025massv,lin2025speculative}, and visual compression~\citep{huang2025specvlm,wang2025flash}.
Concurrent work TIGER~\citep{vo2026tiger} routes a fixed-size Top-$K$ visual subset from the current textual state once per speculative block.
It further trains the drafter with verifier-accepted-prefix rewards, directly aligning the objective with speculative decoding efficiency.
\method{} differs by retrieving a dynamic-cardinality subset for corrected draft states and injecting its readout through cache-compatible hidden-state correction without rewriting historical \kv{} entries.

\paragraph{Visual token selection and compression}
Visual tokens are a major source of prefill cost, attention computation, memory traffic, and \kv{}-cache footprint \cite{huang2025dynamic, tu2025vl} in VLM inference.
Existing acceleration methods therefore reduce the visual sequence through pruning \cite{chen2024image}, parameter-free pooling \cite{yao2024deco}, resampling \cite{alayrac2022flamingo}, convolutional downsampling \cite{cha2024honeybee, chu2023mobilevlm}, token hiding \cite{zhang2025llava}, or query-aware selection \cite{zhu2024focusllava, zhang2024sparsevlm}.
Within multimodal speculative decoding, visual reduction has been used to lower draft latency or transfer cost, such as vision-aware compression~\citep{kang2026vispec,hu2026dream}, elastic or latent-guided compression~\citep{huang2025specvlm,wang2025flash}, hidden visual-token representations~\citep{xie2026hivis}, and visual or \kv{} information pruning~\citep{jia2026covspec,yang2025aasd}.

However, most of these methods choose a compact visual representation before decoding or reuse one visual interface throughout generation.
This fixed-selection assumption is mismatched with speculative decoding, where different steps and branches may require different visual evidence.
\method{} treats visual evidence as a reusable memory: image evidence is stored once, each corrected state selects a state-conditioned subset, and the readout corrects the current scoring representation rather than being inserted as new context tokens.
Thus, \method{} enables state-conditioned visual focus while preserving \kv{}-cache reuse.

\begin{figure}[t]
    \centering
    \includegraphics[width=0.96\columnwidth]{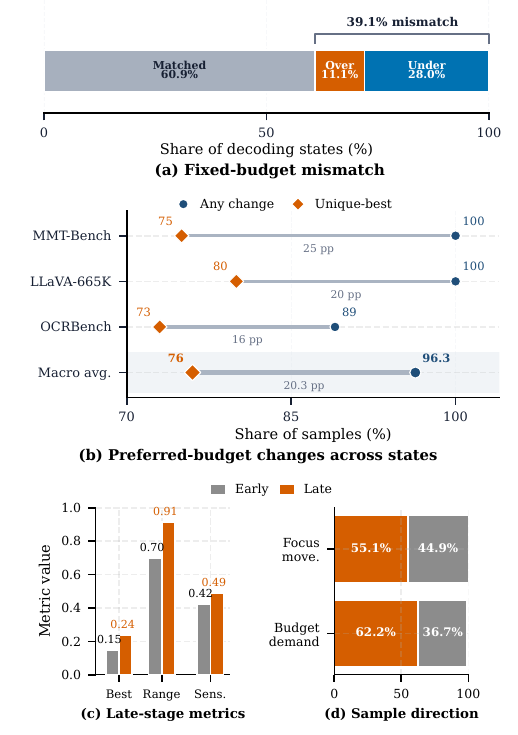}
    \caption{State-dependent visual demand in multimodal speculative decoding.
    (a) Fixed visual budgets cause both under- and over-supply.
    (b) Preferred budgets vary across decoding states.
    (c) Later positions show higher visual demand, accept-length variation, and budget sensitivity.
    (d) Late-stage states exhibit stronger visual-focus shifts.}
    \label{fig:visual-budget-motivation}
\end{figure}

\begin{figure*}[!t]
    \centering
    \includegraphics[width=0.96\linewidth]{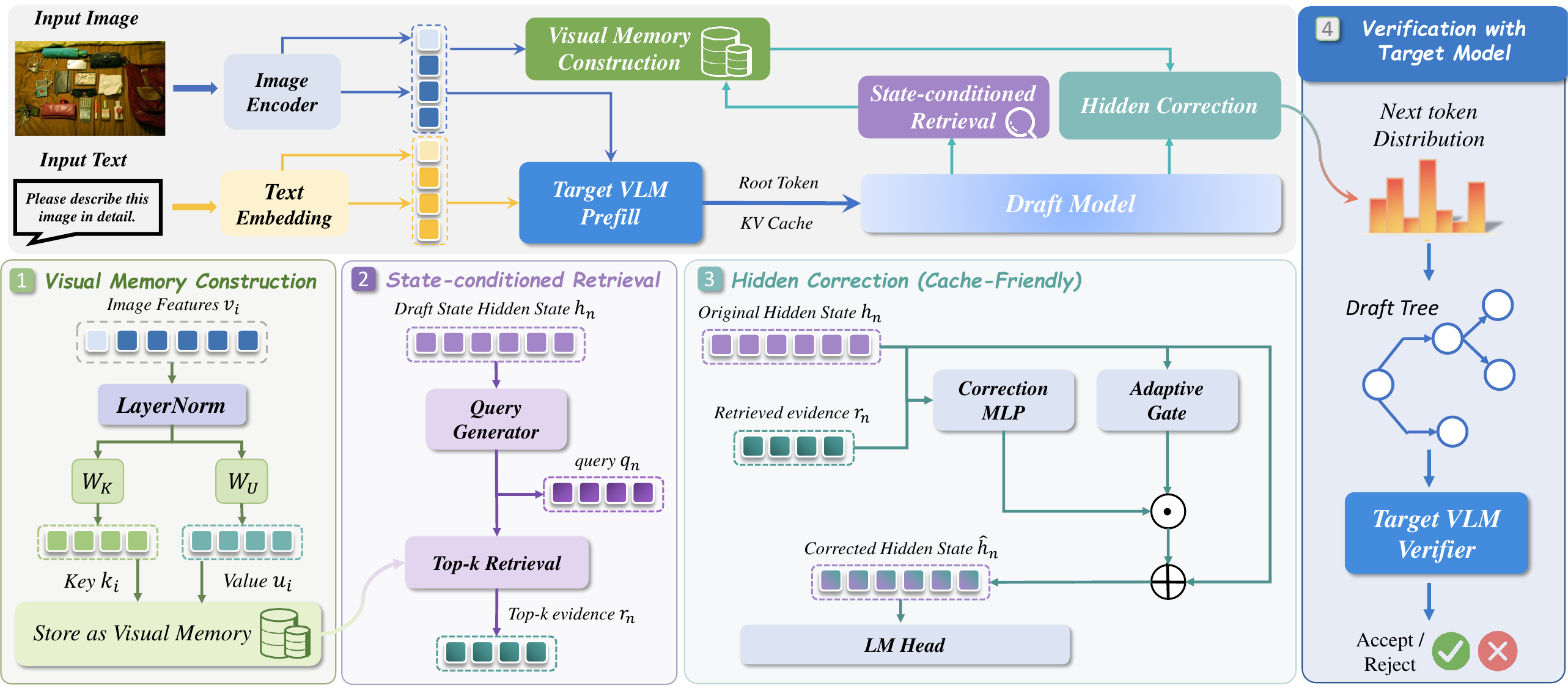}
    \caption{Overview of \method{}.
    The target VLM performs multimodal prefill, while projected visual
    features are cached once as an image-specific key--value memory.
    A draft-tree node uses its hidden state and depth embedding to score
    the memory and attend the top relevant entries. A gated correction
    MLP fuses the retrieved evidence into the current scoring representation
    without modifying the historical \kv{} cache. The corrected distribution
    expands the draft tree, which is then verified in parallel by the target
    VLM.}
    \label{fig:overview}
\end{figure*}



\section{Method}

\paragraph{Problem formulation}
Let $p_T$ and $p_D$ denote the target \vlm{} and lightweight drafter,
respectively. At each speculative step, $p_D$ expands a draft tree whose
candidate nodes are verified in parallel by $p_T$. For a node $n$ with
hidden state $h_n$ and depth $\ell_n\in\{0,\ldots,D\}$, and projected
visual tokens $V=\{v_i\}_{i=1}^{N}$, we define the state-conditioned
visual readout as
\begin{equation}
    r_n = R_{\theta}(h_n,\ell_n,V),
\end{equation}
where $R_{\theta}$ retrieves visual evidence relevant to node $n$.

A fixed visual interface is inadequate because visual-budget sensitivity varies
across states and is non-monotonic. We define the preferred
budget as the smallest tested budget attaining the maximum accept length, rather
than the highest token probability or lowest measured latency. As shown in
Figures~\ref{fig:intro} and~\ref{fig:visual-budget-motivation}, a 25\%
visual budget is optimal for 56.0\% of evaluated states, whereas the
default and full budgets are optimal for only 4.0\% and 2.9\%.
Preferred diagnostic budgets also change across branches and tend to increase at
later positions on OCRBench, motivating state-conditioned evidence selection.

Relevant evidence can increase the average accept length $\tau$, but
unnecessary visual computation raises draft latency. Moreover,
inserting state-specific visual tokens into the autoregressive context
would invalidate historical \kv{} entries. \method{} therefore stores
image evidence in an external memory, selects a dynamic number of entries for
each state, and applies each retrieved readout only to the current scoring
representation. This changes both the selected evidence and retrieval cardinality without changing the cached context.

\paragraph{Image-specific visual memory}
For each input image, the target-side visual encoder and multimodal projector produce the projected visual-token sequence $V=\{v_i\}_{i=1}^{N}$.
\method{} then constructs a per-image visual memory once:
\begin{equation}
    M_{\mathrm{img}}=\{(k_i,u_i,m_i)\}_{i=1}^{N}
\end{equation}
where
\begin{equation}
    k_i = W_K \mathrm{LN}(v_i) \quad
    u_i = W_U \mathrm{LN}(v_i).
\end{equation}
Here, $\mathrm{LN}$ denotes layer normalization. The matrices $W_K\in\mathbb{R}^{d_k\times d_v}$ and
$W_U\in\mathbb{R}^{d_u\times d_v}$ are learned key and value projections, respectively. Thus, $k_i\in\mathbb{R}^{d_k}$ and $u_i\in\mathbb{R}^{d_u}$. The binary mask $m_i\in\{0,1\}$ is one for a valid visual position and zero for a padded position.

The same memory is shared across all speculative iterations and all draft-tree nodes for the image.
By caching this memory, \method{} avoids repeatedly passing the full visual sequence through the draft model while retaining access to fine-grained visual evidence.

\paragraph{State-conditioned top-$K_{r,n}$ visual evidence retrieval}
For each node $n$ at an enabled correction depth, \method{} forms a state-conditioned query from its
hidden state and tree depth, and scores every valid visual-memory key:
\begin{equation}
    q_n=W_Q\bigl(\mathrm{LN}(h_n)+E_{\mathrm{dep}}[\ell_n]\bigr),
    \qquad
    s_{n,i}=\frac{q_nk_i^\top}{\sqrt{d_k}}.
\end{equation}
Let $\mathcal{V}=\{i\mid m_i=1\}$ be the valid memory indices. We normalize
the scores globally as
\begin{equation}
    \widetilde{\alpha}_{n,i}
    =
    \frac{\exp(s_{n,i})}
         {\sum_{j\in\mathcal{V}}\exp(s_{n,j})},
    \qquad i\in\mathcal{V},
\end{equation}
and let $\pi_n(t)$ denote the index of the $t$-th largest normalized score.
The state-dependent retrieval cardinality is
\begin{equation}
\begin{aligned}
    K_{r,n}^{\mathrm{raw}}
    &=
    \min\left\{
        k:\sum_{t=1}^{k}
        \widetilde{\alpha}_{n,\pi_n(t)}\ge\rho
    \right\},\\
    K_{r,n}
    &=
    \min\left\{
        |\mathcal{V}|,\,
        K_{\max},\,
        \max(K_{\min},K_{r,n}^{\mathrm{raw}})
    \right\},
\end{aligned}
\end{equation}
where $\rho=0.8$, $K_{\min}=32$, and $K_{\max}=256$. We retrieve
$\mathcal{I}_n=\{\pi_n(t)\}_{t=1}^{K_{r,n}}$ and renormalize the selected
scores to obtain
\begin{equation}
    \alpha_{n,i}
    =
    \frac{\exp(s_{n,i})}
         {\sum_{j\in\mathcal{I}_n}\exp(s_{n,j})},
    \qquad
    r_n=\sum_{i\in\mathcal{I}_n}\alpha_{n,i}u_i.
\end{equation}
This cumulative-mass rule introduces no additional trainable parameters or
auxiliary loss.
Every valid memory key is still
scored; the dynamic cardinality changes the number of selected and aggregated
values rather than the number of scored keys.

\paragraph{Cache-friendly hidden-vector correction}
The selected visual readout is not inserted as additional context tokens.
Instead, for a draft node \(n\) at an enabled correction depth, \method{} fuses the visual readout $r_n$ with the current hidden state $h_n$ to predict a $d_h$-dimensional correction vector and a scalar gate that is broadcast across the hidden dimension:
\begin{equation}
    c_n = f_{\theta}([h_n;r_n]) \qquad
    g_n = \sigma(g_{\theta}([h_n;r_n]))
\end{equation}
and applies an RMS-normalized gated correction to the current scoring representation:
\begin{equation}
    \hat h_n = \operatorname{RMSNorm}\!\left(h_n + g_n c_n\right),
    \qquad
    z_n = W_{\mathrm{LM}}\hat h_n
    \label{eq:hidden_correction}
\end{equation}
where $W_{\mathrm{LM}}$ is the language-model head. The retrieved
evidence modifies only the representation used for current-token
scoring; the draft token sequence, attention mask, position identifiers,
and historical \kv{} cache remain unchanged. Separately, a stable
context-\kv{} prefix encodes prompt and prefill information and is
shared across all draft-tree nodes.

At inference time, \method{} constructs the image-specific visual memory
once and reuses the draft model's autoregressive cache throughout
generation. During each speculative iteration, nodes at an enabled correction depth
use their hidden states and depth embeddings to retrieve top relevant
visual readouts, apply Equation~\ref{eq:hidden_correction}, and score
candidate tokens for tree expansion. The target \vlm{} then verifies the
draft tree in parallel, after which the accepted tokens and reusable
caches are updated as in standard speculative decoding.

\begin{algorithm}[t]
\caption{\method{}: Dynamic-Cardinality Visual-Evidence Drafting and Verification}
\label{alg:method}
\begin{algorithmic}[1]

\REQUIRE Target VLM $p_T$, drafter $p_D$, visual tokens $V$,
tree depth $D$, draft budget $B$, threshold $\rho$, and bounds
$K_{\min},K_{\max}$
\ENSURE Target-verified sequence $\mathbf{y}$

\AlgComment{Construct the reusable visual memory.}
\STATE
$M_{\mathrm{img}}\gets
\{(W_K\mathrm{LN}(v_i),W_U\mathrm{LN}(v_i),m_i)\}_{i=1}^{N}$, \\
$\mathcal{V}\gets\{i\mid m_i=1\}$.
\STATE Run target prefill and initialize the draft root and reusable caches.

\WHILE{generation has not terminated}
    \STATE Initialize draft tree $\mathcal{T}$ from the current root.
    \FOR{Each node $n$ at an enabled correction depth}
        \STATE Obtain hidden state $h_n$ and depth $\ell_n$.
        \STATE
        $q_n\gets W_Q(\mathrm{LN}(h_n)+E_{\mathrm{dep}}[\ell_n])$, \\
        $s_{n,i}\gets q_nk_i^\top/\sqrt{d_k}$ for $i\in\mathcal{V}$.

        \AlgComment{Select a bounded cumulative-mass subset.}
        \STATE
        $\widetilde{\alpha}_{n,i}\gets
        \exp(s_{n,i})/\sum_{j\in\mathcal{V}}\exp(s_{n,j})$.
        \STATE Let $\pi_n(t)$ be the index of the $t$-th largest
        $\widetilde{\alpha}_{n,i}$.
        \STATE
        $K_{r,n}^{\mathrm{raw}}\gets
        \min\{k:\sum_{t=1}^{k}
        \widetilde{\alpha}_{n,\pi_n(t)}\ge\rho\}$.
        \STATE
        $K_{r,n}\gets\min\{|\mathcal{V}|,K_{\max},
        \max(K_{\min},K_{r,n}^{\mathrm{raw}})\}$,
        $\quad\mathcal{I}_n\gets
        \{\pi_n(t)\}_{t=1}^{K_{r,n}}$.
        \STATE
        $\alpha_{n,i}\gets
        \exp(s_{n,i})/\sum_{j\in\mathcal{I}_n}\exp(s_{n,j})$, \\
        $r_n\gets\sum_{i\in\mathcal{I}_n}\alpha_{n,i}u_i$.

        \AlgComment{Correct only the current scoring representation.}
        \STATE
        $c_n\gets f_{\theta}([h_n;r_n])$,
        $\quad g_n\gets\sigma(g_{\theta}([h_n;r_n]))$.
        \STATE
        $\hat h_n\gets\operatorname{RMSNorm}(h_n+g_nc_n)$,
        $\quad z_n\gets W_{\mathrm{LM}}\hat h_n$.
        \STATE Add candidates from $z_n$ to $\mathcal{T}$ under budget $B$.
    \ENDFOR

    \STATE
    $\mathbf{y}_{\mathrm{acc}}\gets
    \operatorname{Verify}(p_T,\mathcal{T})$.
    \STATE Append $\mathbf{y}_{\mathrm{acc}}$ to $\mathbf{y}$ and update caches.
\ENDWHILE

\RETURN $\mathbf{y}$
\end{algorithmic}
\end{algorithm}

\paragraph{Training objective}
\method{} is trained using precomputed target-model artifacts.
Each training sample contains input token embeddings, target hidden states, supervision masks, and projected visual-memory tokens.
Let $\Omega$ denote the set of supervised positions. For each
$n\in\Omega$, we assign the exponentially decaying loss weight
\begin{equation}
    w_n=\exp\!\left(-\frac{n-1}{\gamma_w}\right),
\end{equation}
where $\gamma_w>0$ controls the decay rate, and define
$Z_\Omega=\sum_{n\in\Omega}w_n$.

For each $n\in\Omega$, let $\hat h_n$ be the corrected draft hidden state, $h_n^T$ the corresponding target-model hidden state, and $\tilde h_n$ and $\tilde h_n^T$ the draft and target intermediate hidden states.
The draft logits are computed as
\begin{equation}
    z_n = W_{\mathrm{LM}}\hat h_n,
\end{equation}
while the target logits are obtained with the same language-model head:
\begin{equation}
    z_n^T = W_{\mathrm{LM}}h_n^T .
\end{equation}
We define the teacher distribution as $p_n^T=\mathrm{softmax}(z_n^T)$, the draft distribution as $p_n^D=\mathrm{softmax}(z_n)$, and the hard teacher token as
$y_n=\arg\max_j z_{n,j}^T$.

Following \cite{hu2026dream}, the hidden-state alignment term matches both final and intermediate representations:
\begin{equation}
    \begin{aligned}
    \mathcal{L}_{\mathrm{hid}}
    &=
    \frac{1}{Z_\Omega}
    \sum_{n\in\Omega} w_n
    \Biggl(
    \beta_h
    \frac{\|\hat h_n-h_n^T\|_2^2}{d_h}
    \\
    &\qquad\qquad
    +
    \beta_{\mathrm{mid}}
    \frac{\|\tilde h_n-\tilde h_n^T\|_2^2}{d_h}
    \Biggr).
    \end{aligned}
\end{equation}

The distribution distillation term matches the teacher next-token distribution:
\begin{equation}
    \begin{aligned}
    \mathcal{L}_{\mathrm{KL}}
    &=
    \frac{1}{Z_\Omega}
    \sum_{n\in\Omega} w_n
    \mathrm{KL}\!\left(p_n^T \,\|\, p_n^D\right) 
    \end{aligned}
\end{equation}

To shape the draft candidate set, let $\mathcal{Y}$ denote the output
vocabulary and let $K_c$ denote the candidate-set width.
For each supervised position $n$, let
\begin{equation}
    \bar z_{n,(1)}
    \ge
    \bar z_{n,(2)}
    \ge
    \cdots
    \ge
    \bar z_{n,(|\mathcal{Y}|-1)}
\end{equation}
denote the competing logits
\begin{equation}
    \{z_{n,j}: j\in\mathcal{Y},\, j\neq y_n\}
\end{equation}
arranged in descending order.
Thus, $\bar z_{n,(K_c)}$ is the $K_c$-th largest competing logit and
defines the candidate-set boundary. We define the teacher--competitor
logit gap as
\begin{equation}
    \gamma_n
    =
    z_{n,y_n}
    -
    \bar z_{n,(K_c)}.
\end{equation}

The token-level objective combines hard-label supervision with a
smooth candidate-boundary margin:
\begin{equation}
    \begin{aligned}
    \mathcal{L}_{\mathrm{tok}}
    &=
    \frac{1}{Z_\Omega}
    \sum_{n\in\Omega} w_n
    \Biggl[
    \beta_{\mathrm{CE}}
    \operatorname{CE}(y_n,p_n^D)
    \\
    &\qquad\qquad
    +
    \beta_{\mathrm{cmp}}
    \log\!\left(1+\exp(m-\gamma_n)\right)
    \Biggr].
    \end{aligned}
\end{equation}

The overall objective consists of hidden-state alignment, distribution distillation, and token-level candidate shaping:
\begin{equation}
    \mathcal{L} =
    \lambda_{\mathrm{hid}} \mathcal{L}_{\mathrm{hid}}
    + \lambda_{\mathrm{KL}} \mathcal{L}_{\mathrm{KL}}
    + \lambda_{\mathrm{tok}} \mathcal{L}_{\mathrm{tok}} .
\end{equation}

\section{Experiment}

\begin{table*}[!t]
  \centering{
  \small
  \setlength{\tabcolsep}{2pt}
  \renewcommand{\arraystretch}{0.96}
  \begin{tabular}{@{}ll*{9}{c}cc@{}}
    \toprule
    \multirow{2}{*}{Model}
      & \multirow{2}{*}{Methods}
      & \multirow{2}{*}{MMT}
      & \multirow{2}{*}{TextVQA}
      & \multirow{2}{*}{MME}
      & \multirow{2}{*}{ChartQA}
      & \multirow{2}{*}{MathVista}
      & \multirow{2}{*}{OCRBench}
      & \multirow{2}{*}{ScienceQA}
      & \multirow{2}{*}{SEED}
      & \multirow{2}{*}{MMSpec}
      & \multicolumn{2}{c}{Avg} \\
    \cmidrule(lr){12-13}
      & & \multicolumn{9}{c}{} & Speedup & $\tau$ \\
    \midrule

    \multirow{9}{*}{\shortstack[l]{LLaVA-Next-7B}}
      \arbaseline
      \methodgroup{Training-free Methods}
      & Lookahead
      & \sr{0.86} & \sr{0.88} & \sr{1.04} & \sr{0.91} & \sr{0.96}
      & \sr{1.02} & \sr{0.81} & \sr{0.61} & \sr{0.72} & \sr{0.87} & 1.641 \\
      & PLD
      & \sr{1.12} & \sr{1.29} & \sr{1.58} & \sr{1.24} & \sr{1.38}
      & \second{1.36} & \sr{1.03} & \sr{0.85} & \sr{1.13} & \sr{1.22} & 1.281 \\
      \methodgroup{Training-based Methods}
      & EAGLE-3
      & \sr{1.62} & \second{1.70} & \second{1.73} & \sr{1.49} & \sr{1.48}
      & \sr{1.12} & \second{1.45} & \sr{1.32} & \sr{1.38} & \sr{1.48} & 2.674 \\
      & Dream
      & \second{1.79} & \sr{1.64} & \sr{1.65} & \second{1.64} & \second{1.65}
      & \sr{0.92} & \best{1.55} & \second{1.39} & \sr{1.36} & \second{1.51} & 3.460 \\
      & Vispec
      & \sr{1.57} & \sr{1.11} & \sr{0.91} & \sr{1.47} & \best{1.67}
      & \sr{1.32} & \sr{1.20} & \sr{0.78} & \second{1.51} & \sr{1.28} & \secondval{3.662} \\
      & \method{}
      & \best{2.13} & \best{1.74} & \best{1.76} & \best{1.68} & \sr{1.57}
      & \best{1.68} & \sr{1.26} & \best{1.80} & \best{1.54} & \best{1.68} & \textbf{3.960} \\
    \midrule

    \multirow{9}{*}{\shortstack[l]{LLaVA-Next-13B}}
      \arbaseline
      \methodgroup{Training-free Methods}
      & Lookahead
      & \sr{0.81} & \sr{0.83} & \sr{1.13} & \sr{0.94} & \sr{1.02}
      & \sr{0.98} & \sr{0.97} & \sr{0.69} & \sr{0.69} & \sr{0.90} & 1.633 \\
      & PLD
      & \sr{0.95} & \sr{1.05} & \sr{1.47} & \sr{1.17} & \sr{1.32}
      & \sr{1.00} & \sr{0.99} & \sr{0.88} & \sr{0.95} & \sr{1.09} & 1.275 \\
      \methodgroup{Training-based Methods}
      & EAGLE-3
      & \second{1.45} & \best{1.69} & \best{1.72} & \sr{1.26} & \sr{1.51}
      & \sr{1.03} & \best{1.52} & \sr{1.37} & \second{1.36} & \sr{1.44} & 2.843 \\
      & MSD
      & \sr{1.39} & \second{1.66} & \sr{1.28} & \second{1.62} & \sr{1.54}
      & \second{1.42} & \sr{1.46} & \second{1.46} & \sr{1.30} & \second{1.46} & 3.490 \\
      & Vispec
      & \sr{1.43} & \sr{0.98} & \sr{0.62} & \sr{1.26} & \second{1.65}
      & \sr{1.18} & \sr{1.35} & \sr{0.85} & \sr{1.35} & \sr{1.19} & \secondval{3.661} \\
      & \method{}
      & \best{1.83} & \second{1.66} & \second{1.69} & \best{1.75} & \best{1.67}
      & \best{1.56} & \second{1.48} & \best{1.83} & \best{1.49} & \best{1.66} & \textbf{3.882} \\
    \midrule

    \multirow{9}{*}{\shortstack[l]{Qwen2.5-VL-7B}}
      \arbaseline
      \methodgroup{Training-free Methods}
      & Lookahead
      & \sr{1.14} & \sr{0.94} & \sr{1.16} & \sr{1.10} & \sr{1.19}
      & \sr{1.10} & \sr{1.41} & \sr{0.95} & \sr{0.98} & \sr{1.11} & 1.425 \\
      & PLD
      & \sr{0.99} & \sr{0.89} & \sr{1.21} & \sr{1.08} & \sr{1.20}
      & \sr{1.14} & \sr{1.12} & \sr{0.84} & \sr{1.05} & \sr{1.06} & 1.150 \\
      \methodgroup{Training-based Methods}
      & EAGLE-3
      & \second{1.48} & \sr{1.18} & \sr{1.14} & \sr{1.41} & \second{1.78}
      & \sr{1.33} & \sr{1.44} & \sr{1.30} & \sr{1.16} & \sr{1.36} & 2.912 \\
      & MSD
      & \sr{1.41} & \second{1.33} & \second{1.30} & \best{1.53} & \sr{1.39}
      & \second{1.54} & \sr{1.34} & \second{1.37} & \second{1.47} & \second{1.41} & 3.452 \\
      & Vispec
      & \sr{1.37} & \sr{0.66} & \sr{0.81} & \sr{1.17} & \sr{1.69}
      & \sr{0.83} & \second{1.46} & \sr{0.76} & \sr{1.27} & \sr{1.11} & \secondval{3.718} \\
      & \method{}
      & \best{1.65} & \best{1.54} & \best{1.48} & \second{1.51} & \best{1.83}
      & \best{1.58} & \best{1.47} & \best{1.68} & \best{1.61} & \best{1.60} & \textbf{3.925} \\
    \bottomrule
  \end{tabular}
  }
  \caption{End-to-end speedup over autoregressive (AR) decoding and average acceptance length ($\tau$). The AR baseline is fixed at $1.00\times$, and $\tau$ is not applicable. Best and second-best speculative-decoding results within each backbone are shown in bold and underlined, respectively.}
  \label{tab:speedup-avg-tau}
\end{table*}

\paragraph{Experimental setup}
We evaluate \method{} with three target backbones:
LLaVA-Next-7B/13B~\citep{liu2024llavanext} and
Qwen2.5-VL-7B~\citep{bai2025qwen25vltechnicalreport}. All three backbone-specific drafters use this same training
set and are optimized for 12 epochs with AdamW
($\beta_1=0.9$, $\beta_2=0.95$).

Following the benchmark-adaptation protocol of
Dream~\citep{hu2026dream}, we train \method{} on 51,671 samples from
LLaVA-v1.5-Mix-665K and 6,000 samples from six task-oriented datasets:
MMT-Bench~\citep{ying2024mmt}, SEED-Bench-2~\citep{li2024seed}, ScienceQA~\citep{saikh2022scienceqa}, MathVista~\citep{lu2024mathvista}, OCRBench~\citep{liu2024ocrbench}, and ChartQA~\citep{masry2022chartqa}.
For MMT-Bench, SEED-Bench-2, MathVista, and OCRBench, we construct
disjoint training and evaluation partitions using
train\_test\_split with random\_state=42, assigning
1,000 samples to training and up to 3,000 remaining samples to
evaluation. For ChartQA and ScienceQA, we randomly select 1,000
examples from their official training splits and evaluate exclusively
on their official evaluation splits. Thus, each benchmark-derived
training subset is disjoint from its corresponding evaluation
manifest. TextVQA~\citep{singh2019towards}, MME~\citep{fu2026mme}, and MMSpec~\citep{shen2026mmspec} contribute no training samples and
are used only for evaluation. We additionally decontaminate
LLaVA-v1.5-Mix-665K against all evaluation
manifests using official sample identifiers and image hashes, obtaining
zero overlap.

The baselines include Lookahead~\citep{zhao2024lookahead}, PLD,
EAGLE-3~\citep{li2026eagle}, and Vispec~\citep{kang2026vispec}.
Dream~\citep{hu2026dream} is evaluated with LLaVA-Next-7B, whereas
MSD~\citep{lin2025speculative} is evaluated with the other two
backbones.

For LLaVA-Next-7B, LLaVA-Next-13B, and Qwen2.5-VL-7B, the FOVEA
drafters use two, three, and two backbone-compatible Transformer blocks,
with approximately 0.621B, 0.962B, and 0.583B trainable parameters,
respectively. Tree-based methods use candidate width $K_c=6$, maximum
depth $D=6$, and a 40-node draft budget. 
Based on the depth-sensitivity analysis in Appendix, visual correction is applied to nodes at draft-tree depth \(\ell_n \le 1\) in all main experiments. We generate at most 200 tokens using greedy decoding and measure
latency on NVIDIA A100 80GB GPUs. The loss weights
$\lambda_{\mathrm{hid}}$, $\lambda_{\mathrm{KL}}$, and
$\lambda_{\mathrm{tok}}$ are set to 1.0, 0.2, and 1.0, respectively.
We report average accept length $\tau$ and speedup, defined as the wall-clock latency ratio
of autoregressive to speculative decoding on the same examples.
Exact target verification preserves the target model's greedy outputs.

\paragraph{Main Results}
Table~\ref{tab:speedup-avg-tau} compares \method{} with representative speculative decoding baselines across three target backbones and nine multimodal benchmarks.
\method{} achieves the strongest average speedup for every backbone, reaching $1.68\times$, $1.66\times$, and $1.60\times$ on LLaVA-Next-7B, LLaVA-Next-13B, and Qwen2.5-VL-7B, respectively.
Compared with the best competing average in each setting, these results correspond to relative improvements of 11.3\%, 13.7\%, and 13.5\%.
The gains are broad rather than benchmark-specific: \method{} ranks first in 21 of the 27 backbone--benchmark combinations, including 7 of 9 benchmarks on LLaVA-7B, 6 of 9 on LLaVA-13B, and 8 of 9 on Qwen2.5-VL-7B.
Moreover, its average speedup remains above $1\times$ in every evaluated combination, indicating robust acceleration across both perception-heavy and reasoning-intensive workloads.

\method{} also obtains the highest average acceptance length for all three backbones, with $\tau$ values of 3.960, 3.882, and 3.925.
These values improve over the strongest competing acceptance lengths by 8.1\%, 6.0\%, and 5.6\%, respectively.
Importantly, longer accepted branches alone do not guarantee greater end-to-end acceleration: Vispec attains relatively high average acceptance lengths of 3.662, 3.661, and 3.718, yet its corresponding average speedups are only $1.28\times$, $1.19\times$, and $1.11\times$.
In contrast, \method{} improves acceptance and wall-clock speed simultaneously.
This combination indicates that its additional accepted tokens do not incur disproportionate draft-side overhead, consistent with the design of constructing visual memory once and accessing it selectively through cache-compatible hidden-state correction.

\paragraph{Ablation and Analysis}
We conduct component and objective ablations on MMT-Bench and OCRBench
with LLaVA-Next-7B under identical training and decoding settings.
Table~\ref{tab:component-objective-ablation} reports retrained
architectural variants that remove visual correction, the depth
embedding, or the correction gate, together with objective variants
that disable hidden-state alignment, distribution distillation, or the
candidate-boundary margin.


\begin{table}[t]
  \centering
  {%
  \small
  \setlength{\tabcolsep}{2pt}
  \begin{tabular}{@{}lcccc@{}}
    \toprule
    \multirow{2}{*}{\textbf{Variant}}
      & \multicolumn{2}{c}{\textbf{MMT-Bench}}
      & \multicolumn{2}{c}{\textbf{OCRBench}} \\
    \cmidrule(lr){2-3}\cmidrule(lr){4-5}
      & Speedup$\uparrow$ & $\tau\uparrow$
      & Speedup$\uparrow$ & $\tau\uparrow$ \\
    \midrule

    \multicolumn{5}{@{}l}{\textit{Architecture Components}} \\

    \rowcolor{methodgroupgray}
    \textbf{Full \method{}}
      & \textbf{2.13} & \textbf{4.52}
      & \textbf{1.68} & \textbf{3.23} \\
    \quad w/o visual correction
      & 1.52 & 3.12 & 1.12 & 2.52 \\
    \quad w/o depth embedding
      & 2.04 & 4.40 & 1.59 & 3.01 \\
    \quad w/o correction gate
      & 1.95 & 4.31 & 1.53 & 2.97 \\

    \midrule
    \multicolumn{5}{@{}l}{\textit{Training Objectives}} \\

    \quad w/o $\mathcal{L}_{\mathrm{hid}}$
      & 1.89 & 4.26 & 1.48 & 2.87 \\
    \quad w/o $\mathcal{L}_{\mathrm{KL}}$
      & 1.45 & 2.87 & 0.97 & 2.12 \\
    \quad w/o candidate-boundary margin
      & 1.75 & 3.85 & 1.42 & 2.81 \\

    \bottomrule
  \end{tabular}
  }%
  \caption{Component and objective ablations on MMT-Bench and
  OCRBench. Higher values are better. Best results are shown in bold.}
  \label{tab:component-objective-ablation}
\end{table}

Visual correction is the dominant architectural component: removing it
reduces the macro-average speedup and acceptance length by 30.7\% and
27.2\%, respectively, whereas removing the depth embedding or correction
gate causes smaller but consistent declines. Among the training signals,
distribution distillation is the most important; without
$\mathcal{L}_{\mathrm{KL}}$, the average speedup and acceptance length
fall to $1.21\times$ and 2.50, with OCRBench becoming slower than
autoregressive decoding at $0.97\times$. Hidden-state alignment and the
candidate-boundary margin provide additional gains.

To further isolate the effect of state-conditioned retrieval, we compare
Full \method{} with five matched controls: Random, Static Image, Query-agnostic, State-only, and Dense Retrieval.
Across MMT-Bench, OCRBench, and ChartQA, all five controls reduce both
macro-average metrics, with absolute decreases of $0.157$--$0.247$ in
speedup and $0.193$--$0.225$ in acceptance length. Detailed
per-benchmark results and repeated-run variability are reported in the
Appendix.

\begin{figure}[t]
    \centering
    \includegraphics[width=0.96\columnwidth]
    {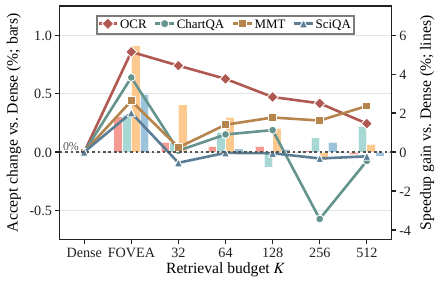}
    \caption{Visual-retrieval budget ablation. FOVEA denotes our retrieval strategy, whereas the remaining setting uses fixed Top-$K$ budgets. Positive values indicate improvements over dense retrieval.}
    \label{fig:visual-budget-ablation}
\end{figure}

\paragraph{State-Conditioned Visual Analysis}
We compare the cumulative-mass retrieval used by \method{}
with dense value aggregation and fixed
$K\in\{32,64,128,256,512\}$.
Figure~\ref{fig:visual-budget-ablation} reports changes in end-to-end
speedup and average acceptance length relative to dense aggregation.

\method{} provides the most balanced cross-task trade-off. It improves
speedup by approximately $5.1\%$ on OCRBench and nearly $3\%$ on both
MMT-Bench and ChartQA. Its acceptance-length changes are modest, reaching
$0.91\%$ on MMT-Bench and $0.50\%$ on ScienceQA, with variations within
approximately $0.25\%$ on ChartQA and OCRBench.
In contrast, fixed budgets exhibit task-dependent and non-monotonic
behavior. These results support adapting the number of aggregated visual values to
each corrected state rather than selecting one fixed cardinality globally.

\begin{figure}[!t]
    \centering
    \includegraphics[width=0.96\columnwidth]{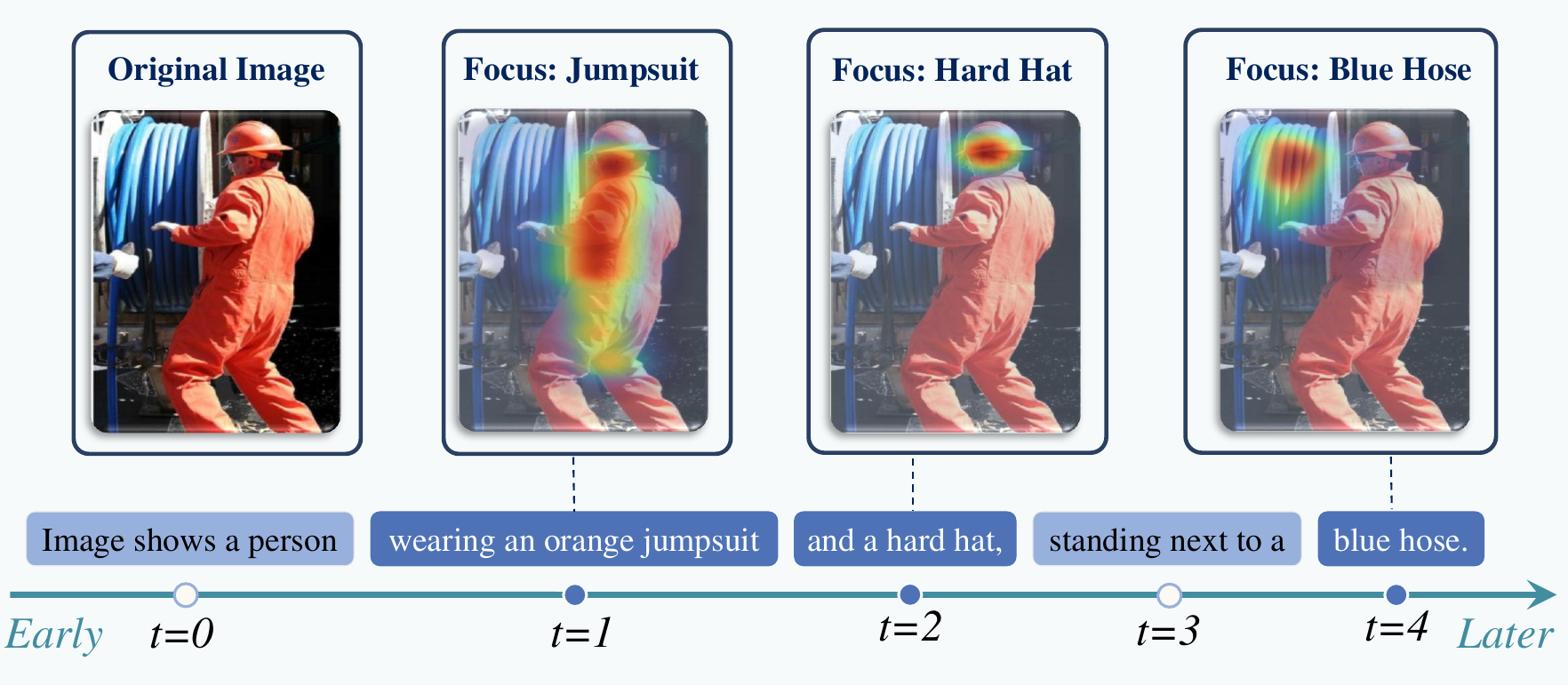}
    \caption{State-conditioned visual retrieval along one decoding
    trajectory. The retrieved focus shifts from the jumpsuit to the
    hard hat and then to the blue hose as generation proceeds.}
    \label{fig:dynamic-visual-focus}
\end{figure}

Figure~\ref{fig:dynamic-visual-focus} provides complementary qualitative
evidence. With the image and visual memory fixed, the retrieved attention
moves from the person's clothing at $t=1$ to the head at $t=2$ and the
left-side hose at $t=4$. The changing focus is therefore driven by the
evolving draft state, confirming that different predictions retrieve
different spatial evidence.

\paragraph{Cache-Friendly Efficiency Analysis}
Table~\ref{tab:cache-efficiency} compares four visual-conditioning
interfaces. \emph{Fixed prefix} reuses static visual conditioning;
\emph{Dynamic insertion} updates the visual prefix and rebuilds its
cache; \emph{Correction + rebuild} applies the same hidden-state
correction as \method{} but disables cache reuse; and
\emph{\method{} correction} modifies only current-token scoring while
preserving all historical \kv{} entries. The latter two variants isolate the latency benefit of cache reuse.


\begin{table}[t]
  \centering
  {\small
  \setlength{\tabcolsep}{1.8pt}
  \begin{tabular}{@{}lcccc@{}}
    \toprule
    \multirow{1}{*}{\textbf{Interface}}
      & \textbf{Time(ms)}$\downarrow$
      & \textbf{$\Delta$Mem.(GiB)}$\downarrow$
      & \textbf{Speedup}$\uparrow$
      & $\boldsymbol{\tau}\uparrow$ \\
    \midrule
    Fixed prefix & \textbf{15.58} & 0.373 & 1.681 & 3.451 \\
    Dynamic insertion & 20.86 & \textbf{0.361} & 1.532 & \textbf{3.719} \\
    Correction + rebuild & 19.90 & 0.377 & 1.764 & 3.647 \\
    \rowcolor{methodgroupgray}
    \textbf{FOVEA correction} & 16.49 & 0.378 & \textbf{1.855} & 3.644 \\
    \bottomrule
  \end{tabular}}
    \caption{Cache-efficiency comparison on MMT-Bench and OCRBench.
    We report draft time, additional peak memory over paired autoregressive
    decoding, end-to-end speedup, and average accept length ($\tau$).
    Best values are shown in bold.}
    \label{tab:cache-efficiency}
\end{table}

Reusing the historical cache reduces draft time from 19.90 to
16.49 ms, a 17.1\% reduction, and increases speedup from
$1.764\times$ to $1.855\times$, while leaving acceptance length nearly
unchanged. Compared with Fixed prefix, \method{}
adds only 0.91 ms per draft iteration but improves acceptance by 5.6\%
and speedup by 10.4\%. Dynamic insertion obtains the highest acceptance
length, yet repeated cache reconstruction raises draft time to
20.86 ms and lowers speedup to $1.532\times$. Peak additional memory
remains similar across all variants, indicating that
cache reconstruction primarily affects latency. 

\section{Conclusion}
Our visual-budget analysis reveals that visual demand varies across decoding
states and that more visual input is not always beneficial. To address this
mismatch, we proposed \method{}, which uses a cumulative-mass
rule to retrieve different amounts and contents of evidence from a reusable
visual memory. A gated correction updates only the current latent
representation, preserving the historical \kv{} cache. Across three backbones
and nine benchmarks, \method{} achieves average speedups of $1.68\times$,
$1.66\times$, and $1.60\times$, with the highest average acceptance
lengths among the evaluated baselines. Further analysis confirms that dynamic
retrieval provides a stronger overall trade-off than fixed-cardinality
alternatives.

\bibliography{aaai2027}

@article{achiam2023gpt,
  title={Gpt-4 technical report},
  author={Achiam, Josh and Adler, Steven and Agarwal, Sandhini and Ahmad, Lama and Akkaya, Ilge and Aleman, Florencia Leoni and Almeida, Diogo and Altenschmidt, Janko and Altman, Sam and Anadkat, Shyamal and others},
  journal={arXiv preprint arXiv:2303.08774},
  year={2023}
}

@article{alayrac2022flamingo,
  title={Flamingo: a visual language model for few-shot learning},
  author={Alayrac, Jean-Baptiste and Donahue, Jeff and Luc, Pauline and Miech, Antoine and Barr, Iain and Hasson, Yana and Lenc, Karel and Mensch, Arthur and Millican, Katherine and Reynolds, Malcolm and others},
  journal={Advances in neural information processing systems},
  volume={35},
  pages={23716--23736},
  year={2022}
}

@inproceedings{cha2024honeybee,
  title={Honeybee: Locality-enhanced projector for multimodal llm},
  author={Cha, Junbum and Kang, Wooyoung and Mun, Jonghwan and Roh, Byungseok},
  booktitle={Proceedings of the IEEE/CVF Conference on Computer Vision and Pattern Recognition},
  pages={13817--13827},
  year={2024}
}

@article{zhu2024focusllava,
  title={Focusllava: A coarse-to-fine approach for efficient and effective visual token compression},
  author={Zhu, Yuke and Xie, Chi and Liang, Shuang and Zheng, Bo and Guo, Sheng},
  journal={arXiv preprint arXiv:2411.14228},
  year={2024}
}

@article{zhang2024sparsevlm,
  title={Sparsevlm: Visual token sparsification for efficient vision-language model inference},
  author={Zhang, Yuan and Fan, Chun-Kai and Ma, Junpeng and Zheng, Wenzhao and Huang, Tao and Cheng, Kuan and Gudovskiy, Denis and Okuno, Tomoyuki and Nakata, Yohei and Keutzer, Kurt and others},
  journal={arXiv preprint arXiv:2410.04417},
  year={2024}
}

@article{chu2023mobilevlm,
  title={Mobilevlm: A fast, strong and open vision language assistant for mobile devices},
  author={Chu, Xiangxiang and Qiao, Limeng and Lin, Xinyang and Xu, Shuang and Yang, Yang and Hu, Yiming and Wei, Fei and Zhang, Xinyu and Zhang, Bo and Wei, Xiaolin and others},
  journal={arXiv preprint arXiv:2312.16886},
  year={2023}
}

@article{wang2024qwen2,
  title={Qwen2-vl: Enhancing vision-language model's perception of the world at any resolution},
  author={Wang, Peng and Bai, Shuai and Tan, Sinan and Wang, Shijie and Fan, Zhihao and Bai, Jinze and Chen, Keqin and Liu, Xuejing and Wang, Jialin and Ge, Wenbin and others},
  journal={arXiv preprint arXiv:2409.12191},
  year={2024}
}

@article{yao2024deco,
  title={Deco: Decoupling token compression from semantic abstraction in multimodal large language models},
  author={Yao, Linli and Li, Lei and Ren, Shuhuai and Wang, Lean and Liu, Yuanxin and Sun, Xu and Hou, Lu},
  journal={arXiv preprint arXiv:2405.20985},
  year={2024}
}

@article{wang2024emu3,
  title={Emu3: Next-token prediction is all you need},
  author={Wang, Xinlong and Zhang, Xiaosong and Luo, Zhengxiong and Sun, Quan and Cui, Yufeng and Wang, Jinsheng and Zhang, Fan and Wang, Yueze and Li, Zhen and Yu, Qiying and others},
  journal={arXiv preprint arXiv:2409.18869},
  year={2024}
}

@inproceedings{zhang2025llava,
  title={Llava-mini: Efficient image and video large multimodal models with one vision token},
  author={Zhang, Shaolei and Fang, Qingkai and Yang, Yang and Feng, Yang},
  booktitle={International Conference on Learning Representations},
  volume={2025},
  pages={53285--53310},
  year={2025}
}

@inproceedings{chen2024image,
  title={An image is worth 1/2 tokens after layer 2: Plug-and-play inference acceleration for large vision-language models},
  author={Chen, Liang and Zhao, Haozhe and Liu, Tianyu and Bai, Shuai and Lin, Junyang and Zhou, Chang and Chang, Baobao},
  booktitle={European Conference on Computer Vision},
  pages={19--35},
  year={2024},
  organization={Springer}
}

@inproceedings{huang2025dynamic,
  title={Dynamic-llava: Efficient multimodal large language models via dynamic vision-language context sparsification},
  author={Huang, Wenxuan and Zhai, Zijie and Shen, Yunhang and Cao, Shaosheng and Zhao, Fei and Xu, Xiangfeng and Ye, Zheyu and Lin, Shaohui},
  booktitle={International Conference on Learning Representations},
  volume={2025},
  pages={69927--69955},
  year={2025}
}

@inproceedings{tu2025vl,
  title={VL-cache: Sparsity and modality-aware KV cache compression for vision-language model inference acceleration},
  author={Tu, Dezhan and Vashchilenko, Danylo and Lu, Yuzhe and Xu, Panpan},
  booktitle={International Conference on Learning Representations},
  volume={2025},
  pages={219--239},
  year={2025}
}

@inproceedings{ku2024findings,
  title={Findings of the Association for Computational Linguistics: ACL 2024},
  author={Ku, Lun-Wei and Martins, Andr{\'e} FT and Srikumar, Vivek},
  booktitle={Findings of the Association for Computational Linguistics: ACL 2024},
  year={2024}
}

@article{touvron2023llama,
  title={Llama: Open and efficient foundation language models},
  author={Touvron, Hugo and Lavril, Thibaut and Izacard, Gautier and Martinet, Xavier and Lachaux, Marie-Anne and Lacroix, Timoth{\'e}e and Rozi{\`e}re, Baptiste and Goyal, Naman and Hambro, Eric and Azhar, Faisal and others},
  journal={arXiv preprint arXiv:2302.13971},
  year={2023}
}

@inproceedings{li2023blip,
  title={Blip-2: Bootstrapping language-image pre-training with frozen image encoders and large language models},
  author={Li, Junnan and Li, Dongxu and Savarese, Silvio and Hoi, Steven},
  booktitle={International conference on machine learning},
  pages={19730--19742},
  year={2023},
  organization={PMLR}
}

@article{chen2023accelerating,
  title={Accelerating large language model decoding with speculative sampling},
  author={Chen, Charlie and Borgeaud, Sebastian and Irving, Geoffrey and Lespiau, Jean-Baptiste and Sifre, Laurent and Jumper, John},
  journal={arXiv preprint arXiv:2302.01318},
  year={2023}
}

@article{liu2023visual,
  title={Visual instruction tuning},
  author={Liu, Haotian and Li, Chunyuan and Wu, Qingyang and Lee, Yong Jae},
  journal={Advances in neural information processing systems},
  volume={36},
  pages={34892--34916},
  year={2023}
}

@article{lu2022learn,
  title={Learn to explain: Multimodal reasoning via thought chains for science question answering},
  author={Lu, Pan and Mishra, Swaroop and Xia, Tanglin and Qiu, Liang and Chang, Kai-Wei and Zhu, Song-Chun and Tafjord, Oyvind and Clark, Peter and Kalyan, Ashwin},
  journal={Advances in neural information processing systems},
  volume={35},
  pages={2507--2521},
  year={2022}
}

@article{li2024eagle,
  title={Eagle: Speculative sampling requires rethinking feature uncertainty},
  author={Li, Yuhui and Wei, Fangyun and Zhang, Chao and Zhang, Hongyang},
  journal={arXiv preprint arXiv:2401.15077},
  year={2024}
}

@inproceedings{li2024eagle2,
  title={Eagle-2: Faster inference of language models with dynamic draft trees},
  author={Li, Yuhui and Wei, Fangyun and Zhang, Chao and Zhang, Hongyang},
  booktitle={Proceedings of the 2024 conference on empirical methods in natural language processing},
  pages={7421--7432},
  year={2024}
}

@article{li2026eagle,
  title={Eagle-3: Scaling up inference acceleration of large language models via training-time test},
  author={Li, Yuhui and Wei, Fangyun and Zhang, Chao and Zhang, Hongyang},
  journal={Advances in Neural Information Processing Systems},
  volume={38},
  pages={136737--136756},
  year={2026}
}

@article{cai2024medusa,
  title={Medusa: Simple llm inference acceleration framework with multiple decoding heads},
  author={Cai, Tianle and Li, Yuhong and Geng, Zhengyang and Peng, Hongwu and Lee, Jason D and Chen, Deming and Dao, Tri},
  journal={arXiv preprint arXiv:2401.10774},
  year={2024}
}

@inproceedings{leviathan2023fast,
  title={Fast inference from transformers via speculative decoding},
  author={Leviathan, Yaniv and Kalman, Matan and Matias, Yossi},
  booktitle={International Conference on Machine Learning},
  pages={19274--19286},
  year={2023},
  organization={PMLR}
}

@article{hu2026dream,
  title={Dream: Drafting with refined target features and entropy-adaptive cross-attention fusion for multimodal speculative decoding},
  author={Hu, Yunhai and Xia, Tianhua and Liu, Zining and Raman, Rahul and Liu, Xingyu and Bao, Bo and Sather, Eric and Thangarasa, Vithursan and Zhang, Sai Qian},
  journal={Advances in Neural Information Processing Systems},
  volume={38},
  pages={167592--167612},
  year={2026}
}

@article{kang2026vispec,
  title={Vispec: Accelerating vision-language models with vision-aware speculative decoding},
  author={Kang, Jialiang and Shu, Han and Li, Wenshuo and Zhai, Yingjie and Chen, Xinghao},
  journal={Advances in Neural Information Processing Systems},
  volume={38},
  pages={115511--115532},
  year={2026}
}

@article{huang2025specvlm,
  title={SpecVLM: Fast Speculative Decoding in Vision-Language Models},
  author={Huang, Haiduo and Yang, Fuwei and Liu, Zhenhua and Yin, Xuanwu and Li, Dong and Ren, Pengju and Barsoum, Emad},
  journal={arXiv preprint arXiv:2509.11815},
  year={2025}
}

@article{chen2026dflash,
  title={DFlash: Block Diffusion for Flash Speculative Decoding},
  author={Chen, Jian and Liang, Yesheng and Liu, Zhijian},
  journal={arXiv preprint arXiv:2602.06036},
  year={2026}
}

@inproceedings{gagrani2024speculative,
  title={On speculative decoding for multimodal large language models},
  author={Gagrani, Mukul and Goel, Raghavv and Jeon, Wonseok and Park, Junyoung and Lee, Mingu and Lott, Christopher},
  booktitle={Proceedings of the IEEE/CVF Conference on Computer Vision and Pattern Recognition},
  pages={8285--8289},
  year={2024}
}

@article{ganesan2025massv,
  title={Massv: Multimodal adaptation and self-data distillation for speculative decoding of vision-language models},
  author={Ganesan, Mugilan and Segal, Shane and Aggarwal, Ankur and Sinnadurai, Nish and Lie, Sean and Thangarasa, Vithursan},
  journal={arXiv preprint arXiv:2505.10526},
  year={2025}
}

@article{lin2025speculative,
  title={Speculative decoding reimagined for multimodal large language models},
  author={Lin, Luxi and Lin, Zhihang and Zeng, Zhanpeng and Ji, Rongrong},
  journal={arXiv preprint arXiv:2505.14260},
  year={2025}
}

@inproceedings{yang2025aasd,
  title={AASD: Accelerate Inference by Aligning Speculative Decoding in Multimodal Large Language Models},
  author={Yang, Chaoqun and Chen, Ran and Zhang, Muyang and Pang, Weiguang and Chen, Yuzhi and Xu, Rongtao and Fu, Kexue and Wang, Changwei and Gao, Longxiang},
  booktitle={2025 62nd ACM/IEEE Design Automation Conference (DAC)},
  pages={1--7},
  year={2025},
  organization={IEEE}
}

@inproceedings{xie2026hivis,
  title={HiViS: hiding visual tokens from the drafter for speculative decoding in vision-language models},
  author={Xie, Zhinan and Wang, Peisong and Qiu, Shuang and Cheng, Jian},
  booktitle={Proceedings of the IEEE/CVF Conference on Computer Vision and Pattern Recognition},
  pages={8952--8961},
  year={2026}
}

@article{wang2025flash,
  title={FLASH: Latent-Aware Semi-Autoregressive Speculative Decoding for Multimodal Tasks},
  author={Wang, Zihua and Li, Ruibo and Du, Haozhe and Zhou, Joey Tianyi and Zhang, Yu and Yang, Xu},
  journal={arXiv preprint arXiv:2505.12728},
  year={2025}
}

@article{jia2026covspec,
  title={CoVSpec: Efficient Device-Edge Co-Inference for Vision-Language Models via Speculative Decoding},
  author={Jia, Yuanyuan and Tang, Shunpu and Yang, Qianqian},
  journal={arXiv preprint arXiv:2605.02218},
  year={2026}
}

@article{ying2024mmt,
  title={Mmt-bench: A comprehensive multimodal benchmark for evaluating large vision-language models towards multitask agi},
  author={Ying, Kaining and Meng, Fanqing and Wang, Jin and Li, Zhiqian and Lin, Han and Yang, Yue and Zhang, Hao and Zhang, Wenbo and Lin, Yuqi and Liu, Shuo and others},
  journal={arXiv preprint arXiv:2404.16006},
  year={2024}
}

@inproceedings{li2024seed,
  title={Seed-bench: Benchmarking multimodal large language models},
  author={Li, Bohao and Ge, Yuying and Ge, Yixiao and Wang, Guangzhi and Wang, Rui and Zhang, Ruimao and Shan, Ying},
  booktitle={Proceedings of the IEEE/CVF Conference on Computer Vision and Pattern Recognition},
  pages={13299--13308},
  year={2024}
}

@article{saikh2022scienceqa,
  title={Scienceqa: A novel resource for question answering on scholarly articles},
  author={Saikh, Tanik and Ghosal, Tirthankar and Mittal, Amish and Ekbal, Asif and Bhattacharyya, Pushpak},
  journal={International Journal on Digital Libraries},
  volume={23},
  number={3},
  pages={289},
  year={2022}
}

@article{liu2024ocrbench,
  title={Ocrbench: on the hidden mystery of ocr in large multimodal models},
  author={Liu, Yuliang and Li, Zhang and Huang, Mingxin and Yang, Biao and Yu, Wenwen and Li, Chunyuan and Yin, Xu-Cheng and Liu, Cheng-Lin and Jin, Lianwen and Bai, Xiang},
  journal={Science China Information Sciences},
  volume={67},
  number={12},
  pages={220102},
  year={2024},
  publisher={Springer}
}

@inproceedings{masry2022chartqa,
  title={Chartqa: A benchmark for question answering about charts with visual and logical reasoning},
  author={Masry, Ahmed and Tan, Jia Qing and Joty, Shafiq and Hoque, Enamul and others},
  booktitle={Findings of the association for computational linguistics: ACL 2022},
  pages={2263--2279},
  year={2022}
}

@inproceedings{liu2026dream,
  title={DREAM-S: Speculative Decoding with Searchable Drafting and Target-Aware Refinement for Multimodal Generation},
  author={Liu, Zining and Hu, Yunhai and Xia, Tianhua and Bao, Bo and Sather, Eric and Thangarasa, Vithursan and Zhang, Sai Qian},
  booktitle={Proceedings of the 64th Annual Meeting of the Association for Computational Linguistics (Volume 1: Long Papers)},
  pages={47031--47045},
  year={2026}
}

@inproceedings{lu2024mathvista,
  title={Mathvista: Evaluating mathematical reasoning of foundation models in visual contexts},
  author={Lu, Pan and Bansal, Hritik and Xia, Tony and Liu, Jiacheng and Li, Chunyuan and Hajishirzi, Hannaneh and Cheng, Hao and Chang, Kai-Wei and Galley, Michel and Gao, Jianfeng},
  booktitle={International Conference on Learning Representations},
  volume={2024},
  pages={23439--23554},
  year={2024}
}

@inproceedings{singh2019towards,
  title={Towards vqa models that can read},
  author={Singh, Amanpreet and Natarajan, Vivek and Shah, Meet and Jiang, Yu and Chen, Xinlei and Batra, Dhruv and Parikh, Devi and Rohrbach, Marcus},
  booktitle={Proceedings of the IEEE/CVF conference on computer vision and pattern recognition},
  pages={8317--8326},
  year={2019}
}

@article{fu2026mme,
  title={Mme: A comprehensive evaluation benchmark for multimodal large language models},
  author={Fu, Chaoyou and Chen, Peixian and Shen, Yunhang and Qin, Yulei and Zhang, Mengdan and Lin, Xu and Yang, Jinrui and Zheng, Xiawu and Li, Ke and Sun, Xing and others},
  journal={Advances in Neural Information Processing Systems},
  volume={38},
  year={2026}
}

@article{shen2026mmspec,
  title={MMSpec: Benchmarking Speculative Decoding for Vision-Language Models},
  author={Shen, Hui and Wang, Xin and Zhang, Ping and Hsieh, Yunta and Han, Qi and Wan, Zhongwei and Zhang, Ziheng and Zhang, Jingxuan and Xiong, Jing and Liu, Ziyuan and others},
  journal={arXiv preprint arXiv:2603.14989},
  year={2026}
}

@misc{bai2025qwen25vltechnicalreport,
      title={Qwen2.5-VL Technical Report}, 
      author={Shuai Bai and Keqin Chen and Xuejing Liu and Jialin Wang and Wenbin Ge and Sibo Song and Kai Dang and Peng Wang and Shijie Wang and Jun Tang and Humen Zhong and Yuanzhi Zhu and Mingkun Yang and Zhaohai Li and Jianqiang Wan and Pengfei Wang and Wei Ding and Zheren Fu and Yiheng Xu and Jiabo Ye and Xi Zhang and Tianbao Xie and Zesen Cheng and Hang Zhang and Zhibo Yang and Haiyang Xu and Junyang Lin},
      year={2025},
      eprint={2502.13923},
      archivePrefix={arXiv},
      primaryClass={cs.CV},
      url={https://arxiv.org/abs/2502.13923}, 
}

@misc{liu2024llavanext,
  title={Llavanext: Improved reasoning, ocr, and world knowledge},
  author={Liu, Haotian and Li, Chunyuan and Li, Yuheng and Li, Bo and Zhang, Yuanhan and Shen, Sheng and Lee, Yong Jae},
  year={2024}
}

@article{vo2026tiger,
  title={TIGER: Text-Conditioned Visual Gated Routing with Acceptance Alignment for Multimodal Speculative Decoding},
  author={Vo, Quynh and Nguyen, Cong-Duy and Srey, Ponhvoan and Tuan, Luu Anh and Nguyen, Thong},
  journal={arXiv preprint arXiv:2607.11131},
  year={2026}
}

@inproceedings{zhao2024lookahead,
  title={Lookahead: An inference acceleration framework for large language model with lossless generation accuracy},
  author={Zhao, Yao and Xie, Zhitian and Liang, Chen and Zhuang, Chenyi and Gu, Jinjie},
  booktitle={Proceedings of the 30th ACM SIGKDD Conference on Knowledge Discovery and Data Mining},
  pages={6344--6355},
  year={2024}
}

@inproceedings{berdoz2026steering,
  title={Steering pretrained drafters during speculative decoding},
  author={Berdoz, Fr{\'e}d{\'e}ric and Rheinboldt, Peer and Wattenhofer, Roger},
  booktitle={Proceedings of the AAAI Conference on Artificial Intelligence},
  volume={40},
  number={36},
  pages={30067--30075},
  year={2026}
}

@inproceedings{do2026adaspec,
  title={AdaSpec: Adaptive Multilingual Speculative Decoding with Self-Synthesized Language-Aware Training and Vocabulary Simplification},
  author={Do, Dinh-Truong and Le, Nguyen-Khang and Nguyen, Le-Minh},
  booktitle={Proceedings of the AAAI Conference on Artificial Intelligence},
  volume={40},
  number={36},
  pages={30530--30538},
  year={2026}
}

@inproceedings{shi2026scaling,
  title={Scaling llm speculative decoding: Non-autoregressive forecasting in large-batch scenarios},
  author={Shi, Luohe and Li, Zuchao and Zhang, Lefei and Qi, Baoyuan and Liu, Guoming and Zhao, Hai},
  booktitle={Proceedings of the AAAI Conference on Artificial Intelligence},
  volume={40},
  number={39},
  pages={32947--32955},
  year={2026}
}











\appendix

\twocolumn[
  \begin{center}
    {\LARGE\bfseries Appendix}
  \end{center}
  \vspace{1em}
]

\section{Visual-Budget Diagnostic Details}
\label{app:visual-budget-diagnostic}

Table~\ref{tab:draft-training-config} summarizes the model-specific training
configurations.

\begin{figure*}[htbp]
    \centering
    \includegraphics[width=\textwidth]{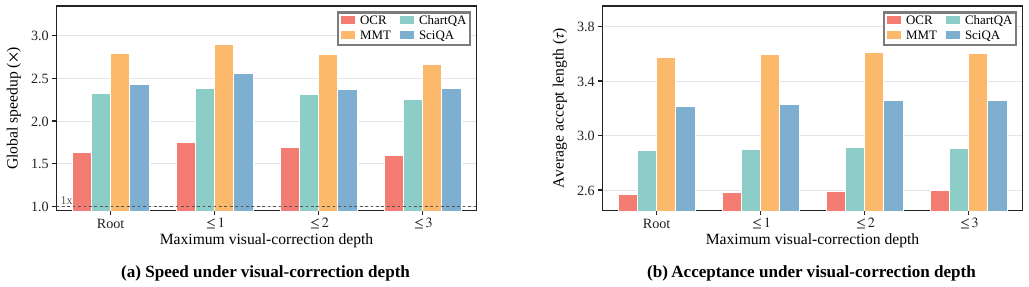}
    \caption{Sensitivity to the visual-correction depth.}
    \label{fig:visual-retrieval-budget}
\end{figure*}

This section provides the complete protocol and metric definitions for the
visual-budget diagnostics summarized in Figures~1 and~2 of the main paper.
The purpose of this analysis is not to
identify a single globally optimal compression ratio, but to test whether the
amount of visual evidence preferred by the drafter remains stable across
decoding states.
This diagnostic varies the amount of visual input to expose the limitations of
one globally fixed interface. FOVEA does not predict these ratios: it uses the
same retrieval cardinality $K_r=16$ at every draft state and adapts which
visual-memory entries are selected.

\subsection{Diagnostic Setup}

\paragraph{Models and data.}
We use LLaVA-NeXT-7B as the target VLM and a draft model trained following the
Dream formulation. We evaluate 1,000 examples from each of
MMT-Bench, LLaVA/665K-Holdout, and OCRBench. Unless stated otherwise, the
decoding configuration is identical to that used in the main experiments:
the drafter uses top-$k=6$, a maximum tree depth of $D=6$, and a budget of 40
candidate nodes. Generation is greedy and produces at most 200 new tokens.

We define a \emph{decoding state} as one complete speculative iteration at a
fixed current prefix. Starting from the same prefix, we separately construct
and verify one draft tree under each visual budget. Thus, a state corresponds
to an entire draft--verify iteration rather than to an individual node within
the draft tree. The frozen diagnostic set contains 4,082 such states.

\paragraph{Visual-budget levels.}
Let $N$ denote the number of projected visual tokens produced by the
LLaVA-NeXT visual encoder and multimodal projector. Because LLaVA-NeXT expands
images according to their resolution, $N$ varies across examples. We evaluate
five visual-budget ratios
\begin{equation}
    \mathcal{R}=\{0,0.25,0.50,0.75,1.00\}.
\end{equation}
For a ratio $r\in\mathcal{R}$, the number of retained visual tokens is
\begin{equation}
    k(r)=
    \begin{cases}
        0, & r=0,\\
        \max(1,\lfloor rN\rfloor), & 0<r<1,\\
        N, & r=1.
    \end{cases}
\end{equation}
The $75\%$ condition is the default visual budget used by the Dream-style
drafter; it is not an additional sixth condition. The $0\%$ and $100\%$
conditions correspond to no visual tokens and full visual access,
respectively.

\begin{table}[t]
    \centering
    \small
    \setlength{\tabcolsep}{4pt}
    \caption{Visual-token ranges in the diagnostic data. The default condition
    retains $\max(1,\lfloor0.75N\rfloor)$ tokens.}
    \label{tab:app-visual-token-ranges}
    \begin{tabular}{@{}lcc@{}}
        \toprule
        Dataset & Original tokens $N$ & Default retained tokens \\
        \midrule
        MMT-Bench            & 1,176--2,928 & 882--2,196 \\
        LLaVA/665K-Holdout   & 1,464--2,352 & 1,098--1,764 \\
        OCRBench             & 870--2,928   & 652--2,196 \\
        \bottomrule
    \end{tabular}
\end{table}

\paragraph{Importance-based token selection.}
The reduced budgets are constructed by importance-based Top-$k$ selection,
rather than uniform sampling, random sampling, pooling, or token merging. For
each current sequence, the target model first performs a forward pass with the
complete visual-token sequence. We take the self-attention map from the last
target-model layer and average over the attention-head and query dimensions to
obtain one importance score for each visual token. For budget $r$, we retain
the $k(r)$ highest-scoring visual tokens. The selected tokens preserve their
original sequence order and remain separate embeddings; all text tokens are
always retained. Consequently, the five conditions differ only in the amount
of visual evidence exposed to the drafter.

\subsection{State-Level Visual-Budget Metrics}

Let $A_s(r)$ be the accept length obtained at decoding state $s$ under visual
budget $r\in\mathcal{R}$. We define the preferred budget as
\begin{equation}
    B_s = \min\!\left(\arg\max_{r\in\mathcal{R}} A_s(r)\right).
    \label{eq:app-preferred-budget}
\end{equation}
The minimum provides a deterministic tie-breaking rule: when multiple budgets
produce the same maximum accept length, we select the smallest budget that
achieves that maximum. This definition measures the least visual evidence
needed to attain the best observed drafting quality at the state.

The resulting distribution is strongly non-uniform. A $25\%$ visual budget is
preferred at $56.0\%$ of the evaluated states, whereas the default $75\%$
budget and full visual access are preferred at only $4.0\%$ and $2.9\%$ of
states, respectively. Together with the variation of $B_s$ across states from
the same example, this result shows that a globally fixed visual budget does
not match the evidence requirements of all speculative iterations. This
diagnostic motivates state-conditioned evidence selection, but FOVEA itself
does not change the retrieval cardinality across states.

\paragraph{Under- and over-supply relative to the default budget.}
To isolate the mismatch between a low visual budget and the default interface,
we compare only the $25\%$ and $75\%$ conditions. The $0\%$, $50\%$, and
$100\%$ conditions do not participate in this particular statistic. We define
\begin{align}
    \text{Under-supplied}(s)
        &\iff A_s(0.25) < A_s(0.75),\\
    \text{Over-supplied}(s)
        &\iff A_s(0.25) > A_s(0.75),\\
    \text{Matched}(s)
        &\iff A_s(0.25) = A_s(0.75).
\end{align}
The low-budget condition is under-supplied when increasing the visual budget
from $25\%$ to the $75\%$ default allows the target model to accept more draft
tokens. Conversely, the default condition is over-supplied when reducing the
budget to $25\%$ increases the accept length. Among the 4,082 states, 1,143
($28.0\%$) are under-supplied, 453 ($11.1\%$) are over-supplied, and 2,486
($60.9\%$) are matched. The presence of both strict inequalities demonstrates
that the relationship between visual budget and drafting quality is
non-monotonic: neither the low budget nor the larger default budget dominates
across states.

\subsection{Visual Demand Along the Decoding Trajectory}

We next examine whether visual demand changes as generation proceeds. This
analysis uses speculative-iteration position rather than the number of emitted
tokens. For a sample containing $I>1$ speculative iterations, the normalized
position of iteration $i$ is
\begin{equation}
    p_i=\frac{i-1}{I-1},
    \qquad
    b_i=\min\!\left(9,\left\lfloor10p_i\right\rfloor\right),
\end{equation}
which assigns the trajectory to ten equal-width bins $b_i\in\{0,\ldots,9\}$.
For the summary in Figure~2(c) of the main paper, we merge these
bins into an Early interval with $p_i<0.6$ and a Late interval with
$p_i\geq0.6$. Thus, Early approximately corresponds to bins 0--5 and Late to
bins 6--9. Each reported quantity is computed directly over all states in the
corresponding interval; we do not first average within each bin and then give
the ten bins equal weight.

In addition to the preferred budget $B_s$ in
Equation~\ref{eq:app-preferred-budget}, we define the state-level accept-length
range and budget-sensitivity indicator as
\begin{align}
    G_s
        &= \max_{r\in\mathcal{R}} A_s(r)
           - \min_{r\in\mathcal{R}} A_s(r),\\
    S_s
        &= \mathbf{1}[G_s>0].
\end{align}
Here, \emph{Best} is the mean of $B_s$, \emph{Range} is the mean of $G_s$, and
\emph{Sensitive} is the mean of $S_s$ over all states in the relevant
interval. Sensitive uses no additional threshold: a state is budget-sensitive
whenever any two tested budgets yield different accept lengths.

On OCRBench, the mean preferred budget increases from 0.150 in the Early
interval to 0.237 in the Late interval. At the same time, the mean
accept-length range grows from 0.699 to 0.915 accepted draft tokens, and the
fraction of budget-sensitive states rises from $42.2\%$ to $48.8\%$. The three
metrics capture complementary effects: later states prefer more visual
evidence on average, their drafting quality varies more strongly across
budgets, and a larger fraction of them respond to the choice of visual budget.

\subsection{Attention-Based Visual-Focus Movement}

Figure~2(d) of the main paper additionally reports an independent
attention-trace diagnostic. This analysis does not measure changes in the
visual-token subset retained by the budget sweep. Instead, at each
autoregressive target-model step $t$, we select the 16 visual tokens with the
highest last-layer attention scores and denote their index set by $F_t$. The
focus movement between adjacent steps is measured by their Jaccard distance,
\begin{equation}
    D_t
    =1-\frac{|F_t\cap F_{t+1}|}{|F_t\cup F_{t+1}|}.
\end{equation}
We assign each adjacent-step pair to Early or Late using the midpoint of its
normalized positions, with midpoints greater than or equal to 0.6 assigned to
Late. For each eligible sample, we compute the mean $D_t$ separately within
the two intervals. A sample exhibits \emph{late-higher focus movement} when
its Late mean is larger than its Early mean.

\begin{table}[t]
  \centering
  \small
  \setlength{\tabcolsep}{7pt}
  \begin{tabular}{@{}lclclc@{}}
    \toprule
    Parameter & Value & Parameter & Value & Parameter & Value \\
    \midrule
    $\gamma_w$               & $1$
    & $\lambda_{\mathrm{hid}}$ & $1.0$
    & $\lambda_{\mathrm{KL}}$  & $0.2$ \\
    $\lambda_{\mathrm{tok}}$ & $1.0$
    & $\beta_h$                & $1.0$
    & $\beta_{\mathrm{mid}}$   & $0.1$ \\
    $\beta_{\mathrm{CE}}$    & $1.5$
    & $\beta_{\mathrm{cmp}}$   & $0.5$
    & $m$                      & $0.5$ \\
    \bottomrule
  \end{tabular}
  \caption{Loss hyperparameters used for training FOVEA.}
  \label{tab:loss-hyperparameters}
\end{table}

We perform an analogous sample-level comparison for visual-budget demand by
comparing the sample's mean preferred budget in the Late and Early intervals.
Among samples with valid measurements in both intervals, 56 of 90 ($62.2\%$)
have a larger mean preferred budget in Late generation, while 54 of 98
($55.1\%$) have larger Late-stage Jaccard distance. These results indicate that
later generation more often exhibits both a higher preferred diagnostic budget
and a change in the attended visual-token set.

The focus-movement diagnostic should be interpreted specifically as a change
in Top-16 token-set membership. It does not account for changes in attention
weights within the selected set, spatial distance between tokens, or the area
of the corresponding image regions. We therefore use it as complementary
evidence of evolving visual focus rather than as a causal attribution measure.

\begin{table*}[!t]
  \centering
  \small
  \setlength{\tabcolsep}{5pt}
  \renewcommand{\arraystretch}{1.05}
  \begin{tabular}{@{}lccccccc@{}}
    \toprule
    Target backbone & Samples & Epochs & Batch size & Learning rate
      & Warmup & GPUs & Wall-clock time \\
    \midrule
    LLaVA-v1.6-Vicuna-7B
      & 57,671 & 12 & $2/4/16$ & $5\times10^{-5}$
      & \shortstack{4,326 / 43,260\\(linear)} & 2 & 32:58:25 \\
    Qwen2.5-VL-7B
      & 57,671 & 12 & $4/1/8$ & $5\times10^{-5}$
      & \shortstack{2,000\\(logarithmic)} & 2 & 32:17:03 \\
    \bottomrule
  \end{tabular}
    \caption{Draft-model training configurations. Batch size is reported as
  per-device batch size / gradient accumulation steps / global batch size.
  The target \vlm{} is frozen in all runs.}
  \label{tab:draft-training-config}
\end{table*}

\begin{table*}[t]
  \centering
  \scriptsize
  \setlength{\tabcolsep}{1.2pt}
  \renewcommand{\arraystretch}{1.08}
  \begin{tabular}{@{}l*{8}{c}@{}}
    \toprule
      & \multicolumn{2}{c}{MMT-Bench}
      & \multicolumn{2}{c}{OCRBench}
      & \multicolumn{2}{c}{ChartQA}
      & \multicolumn{2}{c}{Macro} \\
    \cmidrule(lr){2-3}\cmidrule(lr){4-5}\cmidrule(lr){6-7}\cmidrule(l){8-9}
    Variant
      & $\Delta S$ & $\Delta\tau$
      & $\Delta S$ & $\Delta\tau$
      & $\Delta S$ & $\Delta\tau$
      & $\Delta S$ & $\Delta\tau$ \\
    \midrule
    Random Top-16
      & $+0.377{\pm}0.246$ & $-0.045{\pm}0.120$
      & $-0.209{\pm}0.319$ & $+0.143{\pm}0.097$
      & $-0.922{\pm}0.210$ & $-0.625{\pm}0.519$
      & $-0.176{\pm}0.128$ & $-0.206{\pm}0.071$ \\
    Static Image Top-16
      & $-0.688{\pm}0.063$ & $-0.101{\pm}0.000$
      & $+0.055{\pm}0.041$ & $+1.074{\pm}0.000$
      & $-0.365{\pm}0.019$ & $-1.470{\pm}0.000$
      & $-0.187{\pm}0.018$ & $-0.225{\pm}0.000$ \\
    Query-agnostic Top-16
      & $-0.538{\pm}0.068$ & $-0.101{\pm}0.000$
      & $+0.088{\pm}0.056$ & $+1.076{\pm}0.000$
      & $-0.516{\pm}0.033$ & $-1.428{\pm}0.000$
      & $-0.247{\pm}0.025$ & $-0.213{\pm}0.000$ \\
    State-only Top-16
      & $-0.264{\pm}0.186$ & $-0.264{\pm}0.000$
      & $-0.159{\pm}0.030$ & $+0.037{\pm}0.000$
      & $+0.010{\pm}0.039$ & $-0.130{\pm}0.000$
      & $-0.197{\pm}0.081$ & $-0.204{\pm}0.000$ \\
    Full \method{}
      & $0.000{\pm}0.000$ & $0.000{\pm}0.000$
      & $0.000{\pm}0.000$ & $0.000{\pm}0.000$
      & $0.000{\pm}0.000$ & $0.000{\pm}0.000$
      & $0.000{\pm}0.000$ & $0.000{\pm}0.000$ \\
    Dense Retrieval
      & $-0.174{\pm}0.018$ & $-0.248{\pm}0.000$
      & $+0.050{\pm}0.139$ & $+0.001{\pm}0.000$
      & $-0.161{\pm}0.020$ & $-0.222{\pm}0.000$
      & $-0.157{\pm}0.048$ & $-0.193{\pm}0.000$ \\
    \bottomrule
  \end{tabular}
  \caption{Matched retrieval-control ablation. Each cell reports the absolute
  change from Full \method{} in speedup ($\Delta S$) or average acceptance
  length ($\Delta\tau$); Full \method{} is fixed at zero. Values are mean
  $\pm$ standard deviation over three repeated inference runs using the same
  checkpoint. Higher values are better.}
  \label{tab:matched-retrieval-controls}
\end{table*}

\begin{table*}[t]
  \centering
  \small
  \setlength{\tabcolsep}{5pt}
  \begin{tabular}{@{}lcccc@{}}
    \toprule
    Backbone & $\tilde h_n$ & $h_n$ & $h_n^T$ & $\tilde h_n^T$ \\
    \midrule
    LLaVA-7B
    & Block 1 output & Block 2 output
    & Final target layer & Minimum-entropy layer in $[1,10]$ \\
    LLaVA-13B
    & Block 1 output & Block 3 output
    & Final target layer & Minimum-entropy layer in $[1,10]$ \\
    Qwen2.5-VL-7B
    & Block 1 output & Block 2 output
    & Final target layer & Minimum-entropy layer in $[1,9]$ \\
    \bottomrule
  \end{tabular}
  \caption{Sources of the draft and target states used for supervision.
  Draft states are measured before visual correction.}
  \label{tab:hidden-state-sources}
\end{table*}

\section{Training Setup}

\subsection{Sensitivity to the visual-corretion depth}

We analyze state-conditioned correction from both efficiency and
interpretability perspectives. Using the same checkpoint, we restrict visual correction to
nodes with depth $\ell_n\leq 0$, $1$, $2$, or $3$.
Matched retrieval controls in the Appendix further isolate the
effect of state conditioning. 

As shown in Figure~\ref{fig:visual-retrieval-budget}, extending correction
from the root to depth $\leq 3$ increases the macro-average accept length
from 2.423 to 2.507, but reduces speedup from $1.237\times$ to
$1.190\times$ and the fraction of samples faster than autoregressive
decoding from 75.5\% to 68.2\%. Thus, broader correction provides only
modest acceptance gains while adding draft-side cost, motivating the
lightweight design of the node-specific correction used throughout
\method{}.

\label{app:retrieval-budget}

\subsection{Draft Model Training}
For each target backbone, we train a separate, backbone-specific draft model
while keeping the target \vlm{} frozen throughout training. Training is
performed entirely on the draft side using target artifacts precomputed
offline. Each artifact contains the input token embeddings, target hidden
states and logits, supervision masks, and projected visual tokens required by
the visual-memory module. Consequently, target-side representations serve
only as fixed supervision signals, and no gradient is propagated through the
target \vlm{}.

The draft models for LLaVA-v1.6-Vicuna-7B and Qwen2.5-VL-7B are each trained
on 57,671 target-artifact samples. Of these, 51,671 are obtained from
LLaVA-v1.5-Mix-665K, providing broad multimodal instruction coverage, and the
remaining 6,000 are drawn from six task-oriented sources: MMT-Bench,
SEED-Bench-2, ScienceQA, MathVista, OCRBench, and ChartQA. For the
LLaVA-v1.6-Vicuna-7B run, this benchmark portion contains 1,000 samples from
each source. For the Qwen2.5-VL-7B run, the 6,000 samples are drawn from the
same six-source pool. The visual-memory retrieval cardinality is fixed to
$K_r=16$ during both training and inference and is shared by every draft
state. All valid memory keys are scored before the top-$K_r$ entries are
selected, so state conditioning changes the selected entries rather than the
number of keys scored or values aggregated.

Both draft models are optimized with AdamW using
$\beta_1=0.9$, $\beta_2=0.95$, zero weight decay, gradient clipping at $0.5$,
BF16 mixed precision, DeepSpeed ZeRO-2, and random seed 42. Both use
\texttt{WarmupDecayLR}: the LLaVA-v1.6-Vicuna-7B run applies linear warmup over
4,326 of 43,260 optimizer updates, while the Qwen2.5-VL-7B run uses
DeepSpeed's default logarithmic warmup over 2,000 updates. For the
LLaVA-v1.6-Vicuna-7B run, a per-device batch size of 2, four gradient
accumulation steps, and two GPUs yield a global batch size of 16. For the
Qwen2.5-VL-7B run, a per-device batch size of 4 without gradient accumulation
on two GPUs yields a global batch size of 8. Warmup lengths are counted in
optimizer updates rather than individual microbatches. The frozen
language-model head of the LLaVA-v1.6-Vicuna-7B draft is retained in FP16,
while the remaining trainable computation uses BF16. The reported LLaVA
wall-clock time includes a 100-sample speed evaluation after every epoch.

\subsection{Matched Retrieval-Control Ablation}

Table~\ref{tab:matched-retrieval-controls} shows that all five alternative
interfaces reduce both macro-average metrics relative to Full \method{}. The
absolute macro decreases in speedup and acceptance length
are $(0.176,0.206)$ for Random Top-16, $(0.187,0.225)$ for Static Image
Top-16, $(0.247,0.213)$ for Query-agnostic Top-16, $(0.197,0.204)$ for
State-only Top-16, and $(0.157,0.193)$ for Dense Retrieval. State-only Top-16
is the closest matched selection control, supporting the complementary value
of the depth embedding. Static and query-agnostic retrieval improve OCRBench
but degrade MMT-Bench and ChartQA, indicating a less robust cross-task
trade-off. Random retrieval has the largest variability and the largest
ChartQA speedup decrease, while dense value aggregation remains close to but
below Full \method{} on both macro metrics. Because all repeats use the same
checkpoint, the standard deviations measure repeated-inference variability,
not variation across independent training seeds; zero $\tau$ variation for
deterministic controls is therefore expected.

\section{Implementation and Training Details}
\label{app:implementation-details}

\paragraph{Draft-model architecture.}
Table~\ref{tab:draft-architecture} summarizes the backbone-specific
FOVEA drafters. All reported parameter counts refer to trainable
parameters in the verified final checkpoints. The LLaVA drafters use
standard multi-head attention, whereas the Qwen2.5-VL drafter uses
grouped-query attention with 28 query heads and 4 key--value heads.

\paragraph{Draft and target supervision states.}
For each token position $n$, $\tilde h_n$ is the output of the first
draft Transformer block. The state $h_n$ is the output of the final
draft block before visual correction, corresponding to the second
block for the two-block drafters and the third block for the
LLaVA-13B drafter. The final target state $h_n^T$ is taken from the
last target-model layer.

The intermediate target state $\tilde h_n^T$ is selected independently
for each token. For the LLaVA backbones, we choose the layer with the
lowest attention entropy among target hidden states 1 through 10. For
Qwen2.5-VL-7B, the search covers target hidden states 1 through 9.
This token-adaptive selection supplies the intermediate feature used
by the hidden-state alignment objective.

\paragraph{Visual correction module.}
For every backbone, we set $d_k=d_u=d_h$ and concatenate
$h_n$ and $r_n$ before applying the correction and gate networks:
\begin{equation}
  c_n=f_\theta([h_n;r_n]), \qquad
  g_n=\sigma\!\left(g_\theta([h_n;r_n])\right).
\end{equation}
Both networks contain two biased linear layers, a 512-dimensional
hidden layer, and a SiLU activation, with no dropout. The correction
network outputs a $d_h$-dimensional vector, whereas the gate network
outputs a scalar that is broadcast across the hidden dimension. The
representation passed to the language-model head is
\begin{equation}
  \hat h_n =
  \operatorname{RMSNorm}\!\left(h_n+g_nc_n\right),
  \qquad z_n=W_{\mathrm{LM}}\hat h_n .
\end{equation}

\paragraph{Loss configuration.}
We use the same loss coefficients for all backbone-specific drafters,
as summarized in Table~\ref{tab:loss-hyperparameters}.

\paragraph{Optimization.}
We optimize only the draft-side trainable parameters while keeping
the target VLM frozen. Training uses AdamW with
$\beta_1=0.9$, $\beta_2=0.95$, zero weight decay, gradient clipping
at $0.5$, BF16 mixed precision, DeepSpeed ZeRO-2, and random seed 42.
All drafters are trained for 12 epochs with a peak learning rate of
$5\times10^{-5}$. The per-backbone batch sizes, warmup schedules,
numbers of optimizer updates, GPU counts, and wall-clock training
times are reported in Table~\ref{tab:draft-training-config}.

\end{document}